\documentclass{article}

\usepackage{microtype}
\usepackage{graphicx}
\usepackage{subfigure}
\usepackage{booktabs} % for professional tables
\usepackage{bm}
\usepackage{amsmath}
\usepackage{amsfonts}
\usepackage{multirow}

\DeclareMathOperator{\E}{\mathbb{E}}
\DeclareMathOperator*{\argmax}{argmax}

\def\approxprop{%
  \def\p{%
    \setbox0=\vbox{\hbox{$\propto$}}%
    \ht0=0.6ex \box0 }%
  \def\s{%
    \vbox{\hbox{$\sim$}}%
  }%
  \mathrel{\raisebox{0.7ex}{%
      \mbox{$\underset{\s}{\p}$}%
    }}%
}

\usepackage{hyperref}

\usepackage[accepted]{icml2020}

\icmltitlerunning{Neural Conditional Event Time Models}

\begin{document}

\twocolumn[
\icmltitle{Neural Conditional Event Time Models}



\icmlsetsymbol{equal}{*}

\begin{icmlauthorlist}
\icmlauthor{Matthew Engelhard}{duke}
\icmlauthor{Samuel Berchuck}{duke}
\icmlauthor{Joshua D'Arcy}{duke}
\icmlauthor{Ricardo Henao}{duke}
\end{icmlauthorlist}

\icmlaffiliation{duke}{Duke University}

\icmlcorrespondingauthor{Matthew Engelhard}{m.engelhard@duke.edu}

\icmlkeywords{Survival Analysis, Failure Time, Event Time}

\vskip 0.3in
]
\printAffiliationsAndNotice{}
% leave blank if no need to mention equal contribution

\begin{abstract}
Event time models predict occurrence times of an event of interest based on known features.
Recent work has demonstrated that neural networks achieve state-of-the-art event time predictions in a variety of settings.
However, standard event time models suppose that the event occurs, eventually, in all cases.
Consequently, no distinction is made between $a$) the probability of event occurrence, and $b$) the predicted time of occurrence.
This distinction is critical when predicting medical diagnoses, equipment defects, social media posts, and other events that or may not occur, and for which the features affecting $a$) may be different from those affecting $b$).
In this work, we develop a conditional event time model that distinguishes between these components, implement it as a neural network with a binary stochastic layer representing finite event occurrence, and show how it may be learned from right-censored event times via maximum likelihood estimation.
Results demonstrate superior event occurrence and event time predictions on synthetic data, medical events (MIMIC-III), and social media posts (Reddit), comprising 21 total prediction tasks.
\end{abstract}

\section{Introduction}\label{intro}
The modeling of event times, also known as failure or survival times, is ubiquitous in biostatistics and medicine, economics, operations research, and other fields.
Common approaches include the Cox proportional hazards (Cox-PH) model \cite{cox1972regression}, which assumes the effect of features is multiplicative on the hazard rate, and the accelerated failure time (AFT) model \cite{wei1992accelerated}, in which features accelerate or decelerate the event time density. A key characteristic of event time models, including Cox-PH and AFT, is that they are capable of learning from \textit{censored} event times, particularly right-censored events, wherein the event time is known only to be above a given value. Right-censored events are common in real-world applications, in which events cannot be observed indefinitely.

A number of neural-network-based variations on established event time models have been shown to improve the resulting event time predictions, including several based on Cox-PH \cite{zheng_safe:_2019, katzman_deepsurv:_2018, kvamme_time--event_2019}, and others designed for time-series features \cite{ren_deep_2018, lee2018deephit}. Other neural-network-based models have used alternative loss functions; for example, \citet{chapfuwa2018adversarial} used a nonparametric, adversarially trained model to obtain more accurate event time predictions.
Recent interest in these models reflects the wide range of problems to which they can be applied, and the importance of learning from censored observations rather than discarding them.

However, the standard event time framework, which is shared by the examples above, makes the strong assumption that events of interest will occur, eventually, in all individuals \cite{kalbfleisch2011statistical}.
This assumption, while justified when predicting time of death, for example, limits effectiveness in settings in which censored event times are observed, but events occur only in a subset of the population.
These settings include prediction of medical diagnoses, physical activities, social media activities, interest in specific media content, and many others.
In each case, event time models cannot distinguish between $a$) the probability of event occurrence, and $b$) the time of occurrence, as well as the factors that impact the former versus the latter.

As a motivating example, we consider the problem of medical diagnosis, in which many patients are lost to follow-up, and consequently their subsequent diagnostic status is unknown.
If a binary classifier is chosen to predict diagnosis, the population must be limited to individuals with adequate follow-up, resulting in substantial loss of training data.
Moreover, these individuals may be systematically different from others, leading to biased predictions \cite{von2015completeness}.
If an event time model is applied, on the other hand, factors affecting the time to diagnosis, which include socioeconomic status, racial/ethnic status, and access to care \cite{dovidio2012under}, are conflated with physiologic factors relevant to the underlying condition.

In this work, we address these limitations by formulating a novel conditional event time framework.
Further, we develop a neural conditional event time model in which event occurrences are drawn from a multivariate Bernoulli distribution, \textit{i.e.}, binary stochastic layer, and event times are predicted with a neural accelerated failure time model conditioned on event occurrence.
This approach provides distinct event occurrence and event time predictions, leading to substantially improved prediction performance in both cases.

We evaluate our model on synthetic data, prediction of 10 clinically important events from MIMIC-III \cite{johnson2016mimic}, and prediction of user submissions to popular subreddits, \textit{e.g.}, r/worldnews, from reddit.com, a leading news and web content aggregator.
Model predictions are compared to a standard, \textit{i.e.}, not conditional, neural event time model as well as binary classification of observed event occurrences, with emphasis on predicted event probabilities.
This work, a novel generalization of the event time framework, leverages gradient estimation methods to predict medical conditions, user preferences, and other characteristics not yet observed; and to distinguish the presence of these characteristics from the rate at which they manifest.

\vspace{-3mm}
\section{Related Work}
Neural-network-based (not conditional) event time models have been used to stratify patient risk \cite{ranganath2016deep} and recommend treatment based on electronic health record data and other clinical data \cite{katzman_deepsurv:_2018}, detect online fraud \cite{zheng_safe:_2019}, and predict survival based on blood serum biomarkers \cite{kvamme_time--event_2019}.
Many of these examples are based on Cox-PH \cite{cox1972regression}, but the effect of features is modeled via neural networks \cite{zheng_safe:_2019, katzman_deepsurv:_2018}.
In contrast, \citet{ranganath2016deep} develop a generative model incorporating the Weibull distribution, whereas \citet{chapfuwa2018adversarial} use an adversarial approach to generate nonparametric event time distributions.
Additionally, \citet{ren_deep_2018} use a recurrent neural network to predict event risk based on time-series data, and \citet{lee2018deephit} use a concordance-based loss function that accounts for competing risks, in which only one of several events of interest may occur.

Conditional event time distributions were explored by \citet{elandt-johnson_conditional_1976} and later used in biostatistics to predict long-term versus short-term survival \cite{farewell_use_1982} and oncology outcomes \cite{gaynor_use_1993}.
However, scaling the conditional event time framework to large datasets with many interrelated events of interest requires gradients of the event occurrence model to be backpropagated across a multivariate Bernoulli distribution, {\em i.e.}, binary stochastic layer.
While high-variance gradient estimates can be obtained using the score function estimator \cite{williams1992simple}, a number of lower-variance yet unbiased estimators have been developed more recently \cite{tucker2017rebar, grathwohl2017backpropagation, yin2018arm}.
Alternatively, \citet{jang2016categorical} and \citet{maddison2016concrete} introduce a continuous relaxation of the categorical distribution that results in biased gradient estimates, but allows gradients to be backpropagated directly.

\vspace{-3mm}
\section{Conditional Event Time Models}
Here we introduce the conditional event time (CET) framework, distinguish it from other event time models, and show how conditional event time models may be implemented via neural network with a binary stochastic layer to predict the occurrence of multiple interrelated events on large datasets.

\subsection{Event Time Framework}
Suppose we have $N$ data points in triplets of the form $\mathcal{D} = \{\bm{x}_i, t_i, s_i\}_{i=1}^N$, where the $\bm{x}_i \in \mathbb{R}^d$ are $d$ features associated with individual $i$, the $t_i \in (0, \infty)$ are associated event times, and the $s_i \in \{0, 1\}$ denote whether the $t_i$ are true event times or right-censoring times.
We begin with a single event of interest to simplify our notation, then extend to the more general case in which there are $M$ events of interest.

Let $\mathcal{F}_i \in (0, \infty)$ and $\mathcal{G}_i \in (0, \infty)$ denote random variables associated with events and censoring, respectively, for individual $i$. We suppose the $\mathcal{F}_i$ are drawn independently from event time distribution $f_\theta(t| \bm{x}_i)$, which has associated survivor function $F_\theta(t| \bm{x}_i) = 1 - \int_0^t f_\theta(\tau| \bm{x}_i) d\tau$. Similarly, the $\mathcal{G}_i$ are drawn independently from the unknown censoring density $g_i(t_i)$, which has associated survivor function $G_i(t) = 1 - \int_0^t g_i(\tau) d\tau$.

Our time observations correspond to random variables ${\cal T}_i \in (0, \infty)$ and ${\cal S}_i \in \{0, 1\}$, where ${\cal T}_i = \min(\mathcal{F}_i, \mathcal{G}_i)$ and ${\cal S}_i = \bm{1}({\cal T}_i = \mathcal{F}_i)$ indicates whether ${\cal T}_i$ corresponds to an event time ($s_i=1$) or a censoring time ($s_i=0$).
We suppose the $\mathcal{F}_i$ and $\mathcal{G}_i$ are mutually independent given $\bm{x}_1, ..., \bm{x}_N$, implying that $a$) event times for individuals $i$ and $j$ are conditionally independent given $\bm{x}_i$ and $\bm{x}_j$, and $b$) censoring is \textit{non-informative}, meaning that observing $s_i = 0$ implies only that the event occurred after $t_i$.

The likelihood of observing a particular $\{t_i, s_i\}$ conditioned on features $\bm{x}_i$ is then given by the following:
\begin{align}
\begin{aligned}
    p_\theta(t_i, s_i=1 | \bm{x}_i) = & \ f_\theta(t_i| \bm{x}_i)G_i(t_i) , \\
    p_\theta(t_i, s_i=0 | \bm{x}_i) = & \ g_i(t_i)F_\theta(t_i| \bm{x}_i) , \\
    % \begin{split}
    p_\theta(t_i, s_i | \bm{x}_i) = & \ p_\theta(t_i, s_i=1 | \bm{x}_i)^{s_i}\\
    & \times p_\theta(t_i, s_i=0 | \bm{x}_i)^{1-s_i} .
    % \end{split}
\end{aligned}
\end{align}
Note that the $g_i(\cdot)$ do not depend on $\theta$, therefore parameters $\theta$ of the event time model may be chosen to maximize the likelihood as follows:
\begin{align}
    % \theta_{\rm ML} = \argmax_\theta \ \sum_{i=1}^N & \log p_\theta(t_i, s_i |\bm{x}_i) \notag \\
    % \begin{split}
    \theta_{\rm ML} = \argmax_\theta \ \sum_{i=1}^N & \{ s_i\log f_\theta(t_i| \bm{x}_i) \\
    & + (1-s_i)\log F_\theta(t_i| \bm{x}_i) \} . \notag
    % \end{split}
\end{align}
When there are $M$ events of interest, we suppose the event times $\bm{t}_i \in (0, \infty)^M$ are independent given $\bm{x}_1, ..., \bm{x}_N$, resulting in the following joint density $p_\theta(\bm{t}_i, \bm{s}_i | \bm{x}_i)$:
\begin{equation}
\label{eq:4}
p_\theta(\bm{t}_i, \bm{s}_i | \bm{x}_i) = \prod_{j=1}^M p_\theta^j(t_i^j, s_i^j | \bm{x}_i) .
\end{equation}
The corresponding maximum likelihood estimate is then:
\begin{equation}\label{eq:5}
    \theta_{\rm ML} = \argmax_\theta \ \sum_{i=1}^N\sum_{j=1}^M\log p_\theta^j(t_i^j, s_i^j | \bm{x}_i) .
\end{equation}

For further details, see \citet{kalbfleisch2011statistical}.
Importantly, events $1, \dots, M$ are viewed as \textit{independent} (given $\bm{x}_i$), rather than \textit{competing} events.

\subsection{Conditional Event Time}
\label{cet-framework}
In the conditional event time framework, we are interested in the hidden variable $c_i \in \{0, 1\}$, which indicates whether an event of interest will \textit{ever} occur in individual $i$.
This variable may be viewed as an underlying medical condition, equipment defect, or other characteristic of interest that will eventually manifest given sufficient time.
When $c_i = 1$, the associated event time $\mathcal{F}_i$ is finite, whereas when $c_i=0$, it is not.
As before, we begin with a single event of interest to simplify notation.

Since $t_i$ may not be finite, we augment the domain of $\mathcal{F}_i$ such that $\mathcal{F}_i \in (0, \infty) \cup \{\infty\}$, whereas the censoring time $\mathcal{G}_i$ remains finite.

We would like to have a model $p_\phi(c_i | \bm{x}_i)$, parameterized by $\phi$, for the probability $P(\mathcal{F}_i < \infty | \bm{x}_i)$ that the event will ever occur in individual $i$.
We suppose the $c_i$ depend on $\bm{x}_i$ and follow a Bernoulli distribution:
\begin{equation}
    c_i \mid \bm{x}_i \sim \textrm{Bern}\left(\sigma(h_\phi(\bm{x}_i))\right) ,
\end{equation}
where $\sigma(\cdot)$ denotes the logistic function and $h_\phi(\cdot):\mathbb{R}^d \to \mathbb{R}$ is a function with parameters $\phi$ to be learned along with $\theta$, {\em i.e.}, the parameters of the event time function.

When $c_i=1$, the standard event time model (described previously) applies.
Alternatively, when $c_i=0$ the observed time $t_i$ is guaranteed to be a censoring time, therefore $P(s_i=1|c_i=0,\bm{x}_i) = 0$ and $p(t_i, s_i=1|c_i=0, \bm{x}_i) = 0$ for all $t_i$.
Moreover, since $c_i=0$ implies that $P(\mathcal{G}_i < \mathcal{F}_i) = P(\mathcal{G}_i < \infty) = 1$, the density of ${\cal T}_i$ is simply $g_i(\cdot)$, the density of censoring times.
Consequently, $p_\theta(t_i, s_i | c_i, \bm{x}_i)$ consists of the following four terms:
\begin{align}\label{eq:four}
\begin{aligned}
    p_\theta(t_i, s_i=1 | c_i=1, \bm{x}_i) &= f_\theta(t_i| \bm{x}_i)G_i(t_i) , \\
    p_\theta(t_i, s_i=0 | c_i=1, \bm{x}_i) &= g_i(t_i)F_\theta(t_i| \bm{x}_i) , \\
    p(t_i, s_i=1|c_i=0, \bm{x}_i) &= 0 , \\
    p(t_i, s_i=0|c_i=0, \bm{x}_i) &= g_i(t_i) .
\end{aligned}
\end{align}
In practice, we penalize incorrect prediction of $s_i=1$ when $c_i=0$ by assigning a small probability $0<\epsilon\ll1$ to $P(\mathcal{F}_i < \infty |c_i=0)$, where $\epsilon$ is a hyperparameter of our model tuned on the validation set.
Combining the four terms in~\eqref{eq:four} yields the following expression for $p_\theta(t_i, s_i | c_i, \bm{x}_i)$:
\begin{align}
% \begin{split}
    p_\theta(t_i, s_i | c_i, \bm{x}_i) = & \ p_\theta(t_i, s_i=1 | c_i=1, \bm{x}_i)^{s_ic_i} \\
    & \times p_\theta(t_i, s_i=0 | c_i=1, \bm{x}_i)^{(1-s_i)c_i} \notag \\
    & \times p(t_i, s_i=1 | c_i=0, \bm{x}_i)^{s_i(1-c_i)} \notag \\
    & \times p(t_i, s_i=0 | c_i=1, \bm{x}_i)^{(1-s_i)(1-c_i)} , \notag
% \end{split}
\end{align}
which may be simplified (see Appendix) as follows after removing terms that do not depend on $\theta$ or $c_i$, including $g_i(\cdot)$ and $G_i(\cdot)$:
% note in the appendix that this is an approximation
%
\begin{align}
\begin{aligned}
    p_\theta(t_i, s_i|c_i, \bm{x}_i) \approxprop & \ \epsilon^{s_i(1-c_i)} \\
    & \times f_\theta(t_i|\bm{x}_i)^{s_i}F_\theta(t_i|\bm{x}_i)^{(1-s_i)c_i} .
\end{aligned}
\end{align}
We then use Jensen's inequality to maximize a lower bound on the expected log-likelihood over the latent variables $c_i$:
\begin{align}
% \begin{split}
    & \log p_{\theta, \phi}(\mathcal{D}) = \sum_{i=1}^N\log\E_{c_i \sim p_\phi(c_i|\bm{x}_i)} [p_\theta(t_i, s_i \mid c_i, \bm{x}_i)] \notag \\
    & \hspace{12mm}\ge \sum_{i=1}^N\E_{c_i \sim p_\phi(c_i | \bm{x}_i)} [\log p_\theta(t_i, s_i \mid c_i, \bm{x}_i)] .
% \end{split}
\end{align}
When there are $M$ events of interest, rather than a single one, we suppose the $c_i^j$ are independent given $x_i$ and drawn from a multivariate Bernoulli distribution:
\begin{equation}
    \bm{c}_i \mid \bm{x}_i \sim \prod_{j=1}^M\textrm{Bern}\left(\sigma(h_\phi^j(\bm{x}_i))\right) ,
\end{equation}
where $h_\phi(\cdot): \mathbb{R}^d \to \mathbb{R}^M$ describes the log-odds of all $M$ events.
In parallel with equations \eqref{eq:4} and \eqref{eq:5}, we expand $p_\theta(\bm{t}_i,\bm{s}_i|\bm{c}_i,\bm{x}_i)$ to obtain the following lower bound on the log-likelihood:
\begin{align}
% \begin{split}
    & \log p_{\theta, \phi}(\mathcal{D}) \ge \label{loglik} \\
    & \hspace{16mm} \sum_{i=1}^N \E_{c_i\sim p_\phi(\bm{c}_i | \bm{x}_i)} \left[ \sum_{j=1}^M \log p_\theta^j(t_i^j, s_i^j | \bm{c}_i , \bm{x}_i) \right] . \notag
% \end{split}
\end{align}
Importantly, when $M$ events are present, the fact that one event will (eventually) occur, \textit{i.e.,} $c_i^j = 1$ for some $j$, affects the timing of other events. Thus we have $f_\theta(t_i | \bm{x}_i, \bm{c}_i)$ rather than $f_\theta(t_i | \bm{x}_i)$. This is critical when predicting medical diagnoses, for example, wherein the presence of a given condition may affect health services use or providers' ability to recognize other conditions.
Importantly, however, this dependence requires that $\nabla_\phi\log(p_{\theta, \phi}(\mathcal{D}))$ be backpropagated through samples from a multivariate Bernoulli distribution.

\subsection{Event Occurrence as a Binary Stochastic Layer}

We instantiate $h_\phi(\bm{x}_i)$ and the parameters of the event time distribution $f_\theta(t_i | \bm{x}_i, \bm{c}_i)$ as neural networks, allowing our conditional event time model to be learned via backpropagation.
The form of $f_\theta(\cdot)$ chosen for our experiments is described in the next section, however, the conditional event time framework permits a range of parametric distributions to be used.
Learning the parameters $\theta$ and $\phi$ therefore requires us to calculate both $\nabla_\theta \log p_{\theta, \phi}(\mathcal{D})$ and $\nabla_\phi \log p_{\theta, \phi}(\mathcal{D})$ from equation \eqref{loglik}.
The former may be estimated directly based on samples of $\bm{c}$, but the latter must be backpropagated across these samples, drawn from a multivariate Bernoulli distribution, which is not differentiable.

To estimate $\nabla_\phi \log p_{\theta, \phi}(\mathcal{D})$, we take advantage of recently developed gradient estimators for categorical and Bernoulli random variables.
Specifically, we explore both the Gumbel-Softmax estimator developed concurrently by \citet{jang2016categorical} and \citet{maddison2016concrete}, which is a continuous (and differentiable) relaxation of the categorical distribution; as well as the Augment-Reinforce-Merge (ARM) estimator \cite{yin2018arm}, which provides an unbiased, low-variance gradient estimate for the multivariate Bernoulli distribution specifically.
Although conditional event time models have been proposed in the past, as previously described, these developments allow them to be applied to large datasets containing a large number of features and interrelated event occurrences.
This is critical to their application to the problems we have described, including diagnosis of multiple medical conditions from the electronic health record, and prediction of user interests from social media activity or in recommender systems.

\subsection{Accelerated Failure Time}
We model the event time distribution $f_\theta(\cdot)$ using the accelerated failure time (AFT) model originally proposed by \citet{wei1992accelerated}.
This model supposes that a baseline survival function $F_0(t)$ is scaled uniformly by the effect of features $\bm{x}$ such that $F_\theta(t_i) = F_0(\mu(\bm{x}_i)\cdot t_i)$.
Consequently, the density $f_\theta(t_i|\bm{x}_i)$ may be written as $\mu(\bm{x}_i)f_0(\mu(\bm{x}_i)\cdot t_i)$, and the log-transformed event time random variable $\mathcal{F}_i$ satisfies:
\begin{equation}\label{eq:aft}
    \log (\mathcal{F}_i) = \mu(\bm{x}_i) + \nu_i\varepsilon .
\end{equation}
When $\varepsilon$ is chosen to be normally distributed, {\em i.e.}, $\varepsilon \sim \mathcal{N}(0, 1)$, $f_\theta$ is log-normal with mean and standard deviation given by $\mu(\bm{x}_i)$ and $\nu_i$, respectively.

To account for the dependency of both the scale and uncertainty of event time predictions on $\bm{x}$, we instantiate $\mu(\bm{x}_i)$ and $\nu(\bm{x}_i)$ in~\eqref{eq:aft} using neural networks with parameters $\theta_\mu$ and $\theta_\nu$, respectively, where $\theta = \{\theta_\mu, \theta_\nu\}$, $\mu(\bm{x}_i) = \textrm{NN}(\bm{x}_i; \theta_\mu)$, and $\nu(\bm{x}_i) = \exp{(\textrm{NN}(\bm{x}_i; \theta_\nu))}$.

When predicting $M$ events of interest, we have $\mu(\cdot): \mathbb{R}^{d+M} \to \mathbb{R}^M$ and $\nu(\cdot): \mathbb{R}^{d+M} \to \mathbb{R}^M$, where $\mu^j(\bm{x}_i, \bm{c}_i)$ and $\nu^j(\bm{x}_i, \bm{c}_i)$ specify the parameters of the time distribution $f_\theta^j(t_i^j|\bm{x}_i, \bm{c}_i)$.

This approach provides a simple, flexible event time distribution capable of making accurate event time predictions, as we will show. Having described the conditional event time model, we now present experimental results.

\section{Experiments}
We describe our experimental methods, including performance metrics, baseline models, datasets, and training and evaluation procedures.
We perform experiments on one synthetic and two real-world datasets, comprising a total of 21 distinct prediction tasks.

\subsection{Performance Metrics}

\paragraph{AUC}
The area under the receiver operating characteristic (AUC) assesses binary classification performance of the learned $p_\phi(c | x)$ in predicting whether events of interest will ever occur.
It is calculated using standard methods based on the predicted $p_\phi(c|x)$ and true $c$, on the test set.

\paragraph{Mean Relative Absolute Error (MRAE)}
The accuracy of event time predictions was assessed on the test set by normalizing the absolute error of predictions by the event range, {\em i.e.}, $| t - \hat{t} | / t_\textrm{max}$, where $\hat{t}$ is the predicted event time.
For censored events, predictions are penalized only if the predicted time is before the censoring time, therefore the relative absolute error is defined as $\max(0, t - \hat{t}) / t_\textrm{max}$.

\paragraph{Concordance Index (CI)}
Correct ordering of event time predictions was assessed using the concordance index (CI) developed by \citet{harrell1984regression}, which quantifies the degree to which the order of predicted event times is consistent with the true event times.
Pairs of event times contribute to the CI only if $a$) both event times are known, or $b$) one event time is known, the other is censored, and the known event time occurs before the censoring time.

\subsection{Baseline Models}
We compare the performance of our neural conditional event time model (CET) to $a$) a neural event time model (ET), and $b$) a binary classifier (BC) trained to predict whether events are observed, {\em i.e.}, $s$.
These represent the available alternatives to CET.
All three performance metrics are evaluated on the ET models, but only the AUC can be evaluated on the binary classifier, which does not predict event times.
The ET model matches the baseline model used in \citet{chapfuwa2018adversarial} and is similar to the deep survival models used by \citet{katzman2018deepsurv} and \citet{kvamme2019time}, but we use the accelerated failure time model from CET rather than a Cox proportional hazards framework \cite{cox1972regression}.

Our aim is to evaluate differences between CET, ET, and BC rather than the impact of specific neural network architectures or hyperparameters, therefore, all neural network layers and model hyperparameters are identical between the CET model and the two baselines.
Thus, the ET model $p_{\theta_{\rm ET}}(\bm{t}, \bm{s} | \bm{x})$ matches the event time component $p_{\theta_{\rm CET}}(\bm{t}, \bm{s} | \bm{c}, \bm{x})$ of CET with the exception of the additional input $\bm{c}$, and the BC model matches $p_\phi(\bm{c}|\bm{x})$ from CET.

\subsection{Datasets}
Here we describe the three datasets used in our experiments.
Experimental results are presented in the next section.

\subsubsection{Synthetic}
To illustrate the advantage of the CET model over alternative approaches when learning from censored data, we construct a simple, synthetic dataset with five features and two events of interest.
The eventual occurrence of the first event depends only on the first two features, as shown in the top left panel of Figure \ref{fig:prediction-performance}, whereas the eventual occurrence of the second event depends only on the second two features, as shown in the bottom left panel of Figure \ref{fig:prediction-performance}.
The timing of both events (expected log-time), however, depends linearly on a fifth feature drawn from a standard normal distribution. 
Training, validation, and test sets contain 24k, 8k, and 8k samples, respectively.
Censoring times are uniformly distributed over the full range of event times.

\subsubsection{MIMIC-III}
\label{mimic-methods}
MIMIC-III (Medical Information Mart for Intensive Care), is a de-identified, accessible dataset of intensive care unit stays at the Beth Israel Deaconess Medical Center between 2001 and 2012 \cite{johnson2016mimic}.
With this dataset, we aim to predict whether and when each of 10 important but non-routine laboratory measurements will be collected for the first time based on physiologic and other measurements from the first 24 hours.
Laboratory measurements were selected among those rarely observed in the first 24 hours based on our assessment of their diagnostic and clinical relevance.
For example, observing a ``WBC, CSF'' measurement suggests that a lumbar puncture has been performed.
All 10 laboratory measurements and their rates of occurrence among MIMIC-III stays are presented in Table 1.

\begin{table}[t!]
\caption{Relevance and occurrence rates for MIMIC-III events.}
\label{mimic-stats}
\begin{center}
\begin{small}
\begin{sc}
\begin{tabular}{llr}
\toprule
Lab Measure & Relevance & Rate\\
\midrule
WBC, CSF & Lumbar Puncture & 4.4\%\\
Troponin T & Heart Damage & 35.2\%\\
Intubated & Intubate Patient & 38.5\%\\
WBC, Pleural & Pleural Fluid & 2.8\%\\
TSH & Thyroid Function & 20.5\%\\
D-Dimer & Thromboses & 5.1\%\\
Urobilinogen & Urinalysis & 54.8\%\\
ANA & Autoimmune & 1.8\%\\
Ammonia & Liver Function & 3.6\%\\
Lipase & Pancreatic Func. & 34.8\%\\
\bottomrule
\end{tabular}
\end{sc}
\end{small}
\end{center}
\vspace{-6mm}
\end{table}

The most common chart events (80 total), lab measurements (30 total), and output events (10 total) occurring within the first 24 hours of admission among all stays in the training set were used as features for the prediction tasks.
We ensured that lab measurements selected as events were excluded, but these measurements were not among the 30 most common and were typically observed beyond 24 hours.
All measurements were aggregated by patient by taking the sum and count of all output events; the mean, minimum, maximum, and count of other numeric measurements; and the count of all categorical measurements, resulting in 346 total features.

Event times were censored uniformly over the interval $(0, 2 \cdot t_\text{median}^j)$, where $t_\text{median}^j$ is the median event time at which measurement $j$ was first collected. Note that artificial censoring is critical to our performance evaluation, which requires ground truth event occurrence labels that are distinct from observed events in the training data. MIMIC-III was chosen for its completeness, which allows this ground truth to be determined. In contrast, CET is designed to be effective on datasets with many censored events.

\subsubsection{Reddit}
Reddit is a web content aggregator and discussion forum with approximately 330 million users as of April 2018 \cite{reddit_wired}.
With this dataset, we aim to predict whether and when users will post to each of 9 different subreddits for the first time based on their prior comment history.
Subreddits were hand-selected among those with at least 100k subscribed Reddit users, and all data were collected using the pushshift.io API.
Submission histories prior to Jan 2020 were collected and grouped by user, and individual comment histories from June 2005 to Nov 2017 were collected for all users that posted to at least one of the 9 subreddits.
Users with 20 or more comments prior to their first submission to any of the 9 subreddits were included in the final dataset, which included 492,059 total Reddit users.
The number of total subscribers to each subreddit and the proportion of our sample who posted to it are presented in Table 2.

\begin{table}[h]
\caption{Popularity and submission rate for each Subreddit.}
\label{reddit-stats}
\begin{center}
\begin{small}
\begin{sc}
\begin{tabular}{lrr}
\toprule
Subreddit & Total Subscribed & Rate\\
\midrule
ADHD & 613k & 6.3\%\\
Anxiety & 325k & 8.4\%\\
Books & 17.5m & 13.4\%\\
Depression & 597k & 20.0\%\\
Fitness & 7.7m & 34.6\%\\
LifeProTips & 17.2m & 25.9\%\\
Mental Health & 144k & 2.8\%\\
Suicide Watch & 180k & 7.3\%\\
World News & 23.1m & 0.3\%\\
\bottomrule
\end{tabular}
\end{sc}
\end{small}
\end{center}
\vspace{-4mm}
\end{table}

For the prediction tasks, the first 20 comments from each Reddit user were encoded using Google's Universal Sentence Encoder \cite{cer2018universal}.
Embedded comments were refined via a single fully-connected layer with tanh activation, then aggregated via max and average pooling \cite{shen2018baseline}.
The average time between comments and average comment length (batch-normalized) were used as additional features.
Submission times were censored uniformly over the interval $(0, 2 \cdot t_\text{median}^j)$, where $t_\text{median}^j$ is the median submission time to subreddit $j$. Similar to MIMIC-III (\ref{mimic-methods}), this provides ground truth event occurrence labels that are distinct from observed events in the training data.

\subsection{Training and Evaluation}
For all tasks, data were partitioned into training (60\%), validation (20\%), and test (20\%) sets.
Our aim is to illustrate differences between CET and alternative approaches, therefore we utilize simple multilayer perceptron architectures with a single hidden layer (ReLU activations) for the functions $h_\phi(\cdot)$, $\mu(\cdot)$, and $\nu(\cdot)$. 
All hyperparameters including hidden layer width, Gumbel-Softmax temperature, number of $c_i$ samples, Gumbel-Softmax versus ARM estimator, and the penalty $\epsilon$ (see \ref{cet-framework}) were tuned to maximize AUC on the validation set.
Hyperparameters were then fixed, and all models (CET and baselines) were evaluated 10 times on the test set.
Reported performance measures are the mean and standard deviation of each measure over all 10 runs.
All models were implemented in Tensorflow 1.10 \cite{abadi2016tensorflow} and trained via backpropagation with the Adam optimizer \cite{kingma2014adam} and a batch size of 400, learning rate of $3\times10^{-4}$, and dropout rate of $0.5$.

\section{Experimental Results}
\begin{figure}[t!]
\begin{center}
\centerline{\includegraphics[width=\columnwidth]{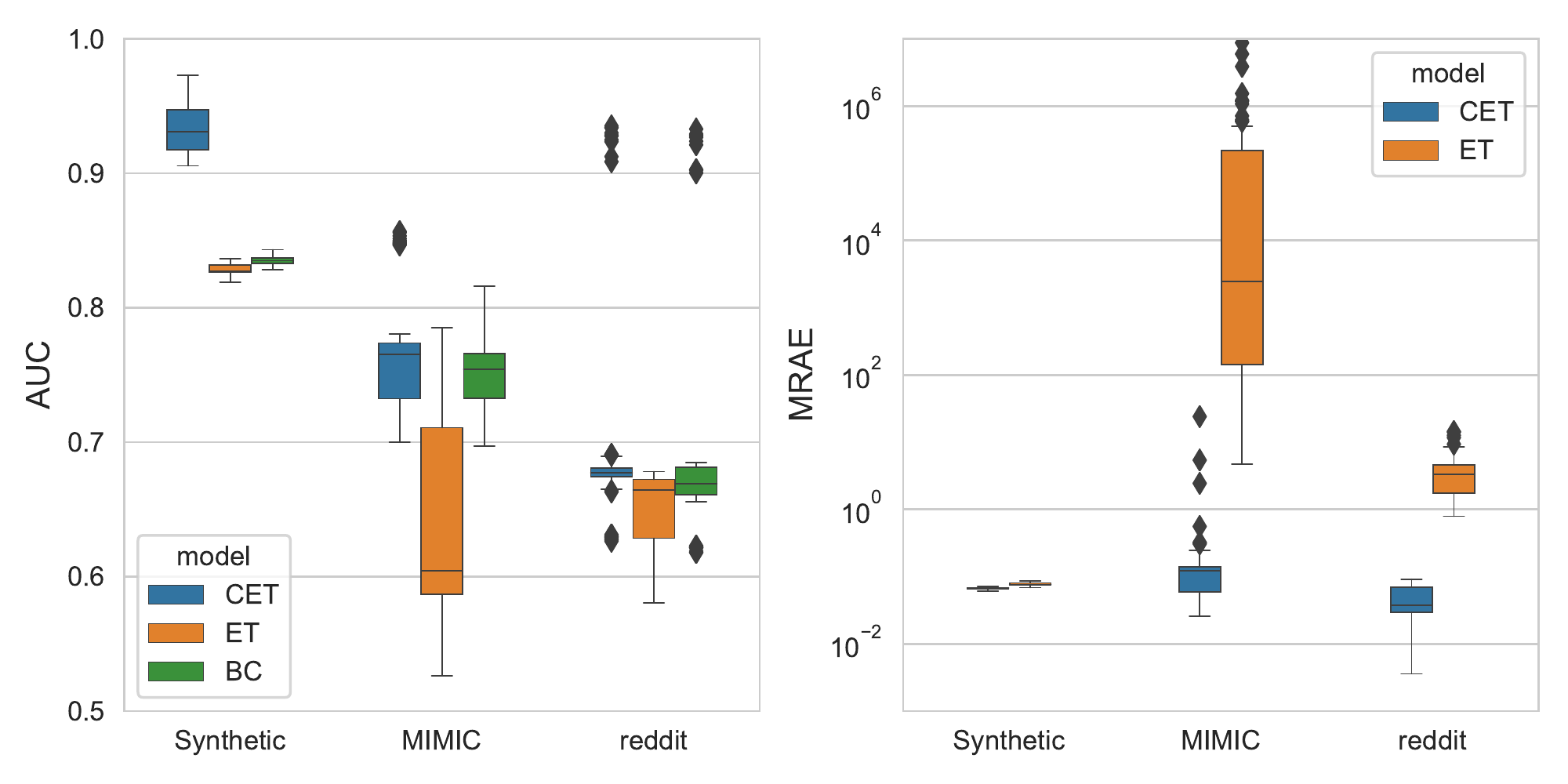}}
\caption{AUC (left) and MRAE (right) for all prediction tasks.}
\label{fig:performance_metrics}
\end{center}
\vskip -0.3in
\end{figure}

Prediction performance (AUC, MRAE) aggregated across all tasks in each dataset is shown in Figure \ref{fig:performance_metrics}. Results show that CET effectively predicts event occurrence despite learning from censored events, with superior performance (AUC, MRAE) compared to ET and BC.

\subsection{Synthetic}
\begin{figure*}[t!]
\begin{center}
\centerline{\includegraphics[width=.8\textwidth]{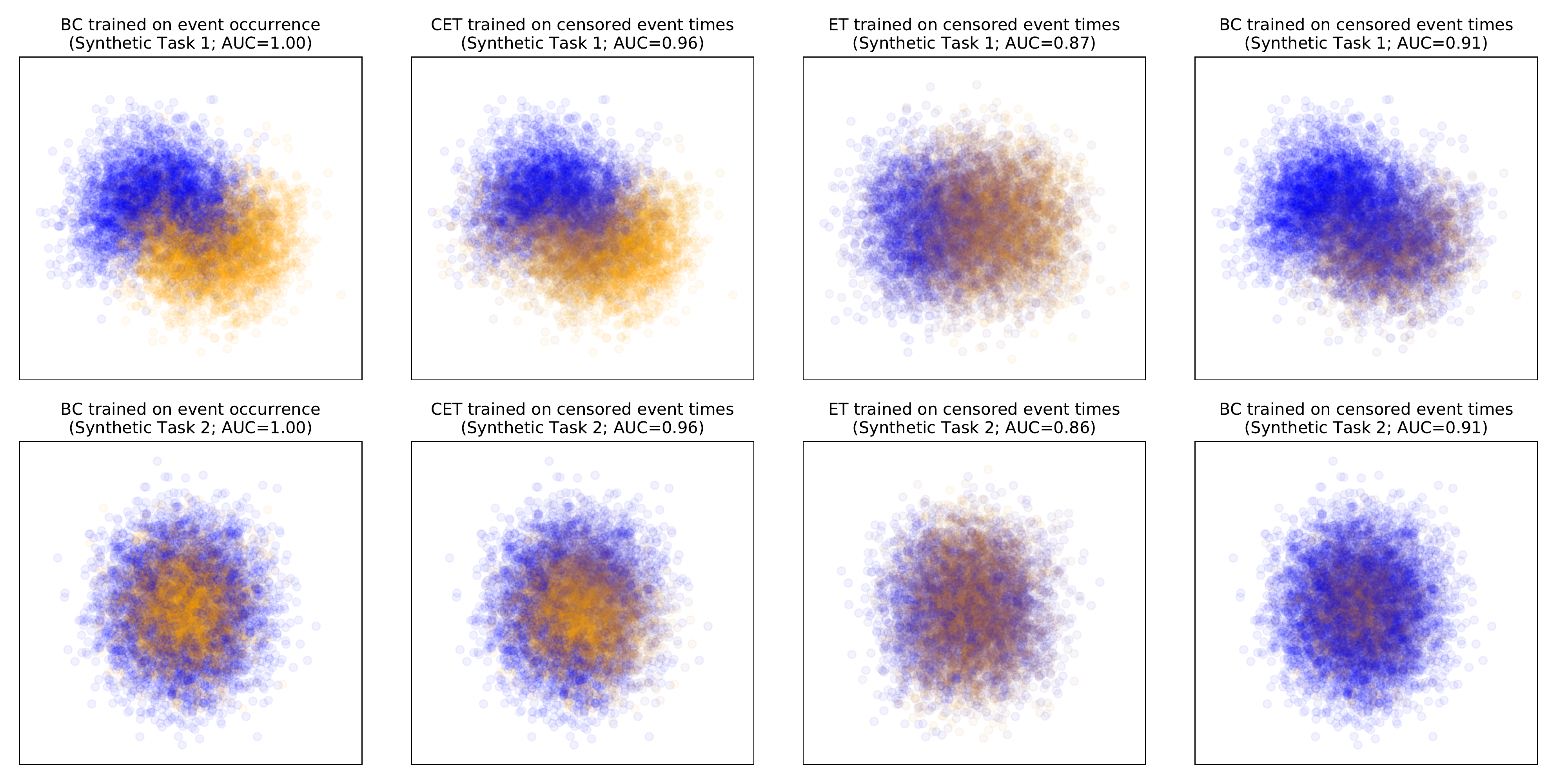}}
\caption{Prediction of event occurrence on synthetic data.}
\label{fig:prediction-performance}
\end{center}
\vskip -0.3in
\end{figure*}

Results on our synthetic dataset (see Table 3), illustrate superior performance of the CET framework compared to existing baselines in $a$) predicting the probability of event occurrence, and $b$) making accurate event time predictions despite using a simple, parametric event time model.
The left panels of Figure~\ref{fig:prediction-performance} show that a simple multilayer perceptron classifier trained directly on known event occurrences (\textit{i.e.,} $\bm{c}$) effectively separates (AUC$\approx$1) individuals in whom the event does versus does not occur in both tasks.
Importantly, this information is not available to the CET and baseline models, which are trained on censored event times.
The middle left panels show that CET also separates these groups effectively despite learning from censored event times only.
In contrast, BC (right panels) cannot distinguish between cases that have been censored and cases in which the event never occurs.
Similarly, although the ET model is able to learn from censored events, it conflates low event probabilities with high event times, leading to poor classification performance (middle right panels).

Figure \ref{fig:prediction-performance} was generated with a lower noise setting compared to the quantitative results, providing clearer separation between groups that allows classification performance to be visualized more effectively.

\begin{table}[h!]
\caption{Performance metrics on synthetic data.}
\label{synthetic-performance}
\begin{center}
\begin{small}
\begin{sc}
\begin{tabular}{llrrr}
\toprule
& Task & AUC & MRAE & CI\\
\midrule
\multirow{3}{*}{CET} & T1 & \textbf{0.93}$\pm$\textbf{0.02} & \textbf{0.07}$\pm$\textbf{0.00} & 0.88$\pm$0.01\\
& T2 & \textbf{0.94}$\pm$\textbf{0.02} & \textbf{0.07}$\pm$\textbf{0.00} & 0.89$\pm$0.00\\
& Avg & \textbf{0.93}$\pm$\textbf{0.02} & \textbf{0.07}$\pm$\textbf{0.00} & 0.88$\pm$0.00\\
\midrule
\multirow{3}{*}{ET} & T1 & 0.83$\pm$0.00 & 0.08$\pm$0.00 & \textbf{0.90$\pm$0.00}\\
& T2 & 0.83$\pm$0.00 & 0.08$\pm$0.00 & \textbf{0.90}$\pm$\textbf{0.00}\\
& Avg & 0.83$\pm$0.00 & 0.08$\pm$0.00 & \textbf{0.90}$\pm$\textbf{0.00}\\
\midrule
\multirow{3}{*}{BC} & T1 & 0.84$\pm$0.00 &  &\\
& T2 & 0.83$\pm$0.00& &\\
& Avg & 0.84$\pm$0.00 & &\\
\bottomrule
\end{tabular}
\end{sc}
\end{small}
\end{center}
\vspace{-3mm}
\end{table}

Compared to the ET model, CET also makes substantially more accurate event time predictions, as shown in Table 3.
This results from the fact that ET must predict a high event time, rather than a low event probability, for individuals in whom the event is not likely to occur.
Consequently, when events do occur in these individuals, the event times predicted by ET are highly inaccurate.
In contrast, CET distinguishes between event probabilities and event times, allowing it to maintain accurate predictions in these cases.

The CI is similar between the CET and ET models, but consistently higher for ET.
This suggests that the ET model is more effective in correctly ordering observed, {\em i.e.}, non-censored, events.
These results are consistent with the fact that the ET objective is designed solely to optimize this ordering, whereas the CET objective also seeks to optimize the predicted probability of event occurrence.

\begin{figure}[h]
\begin{center}
\vspace{-2mm}
\centerline{\includegraphics[width=\columnwidth]{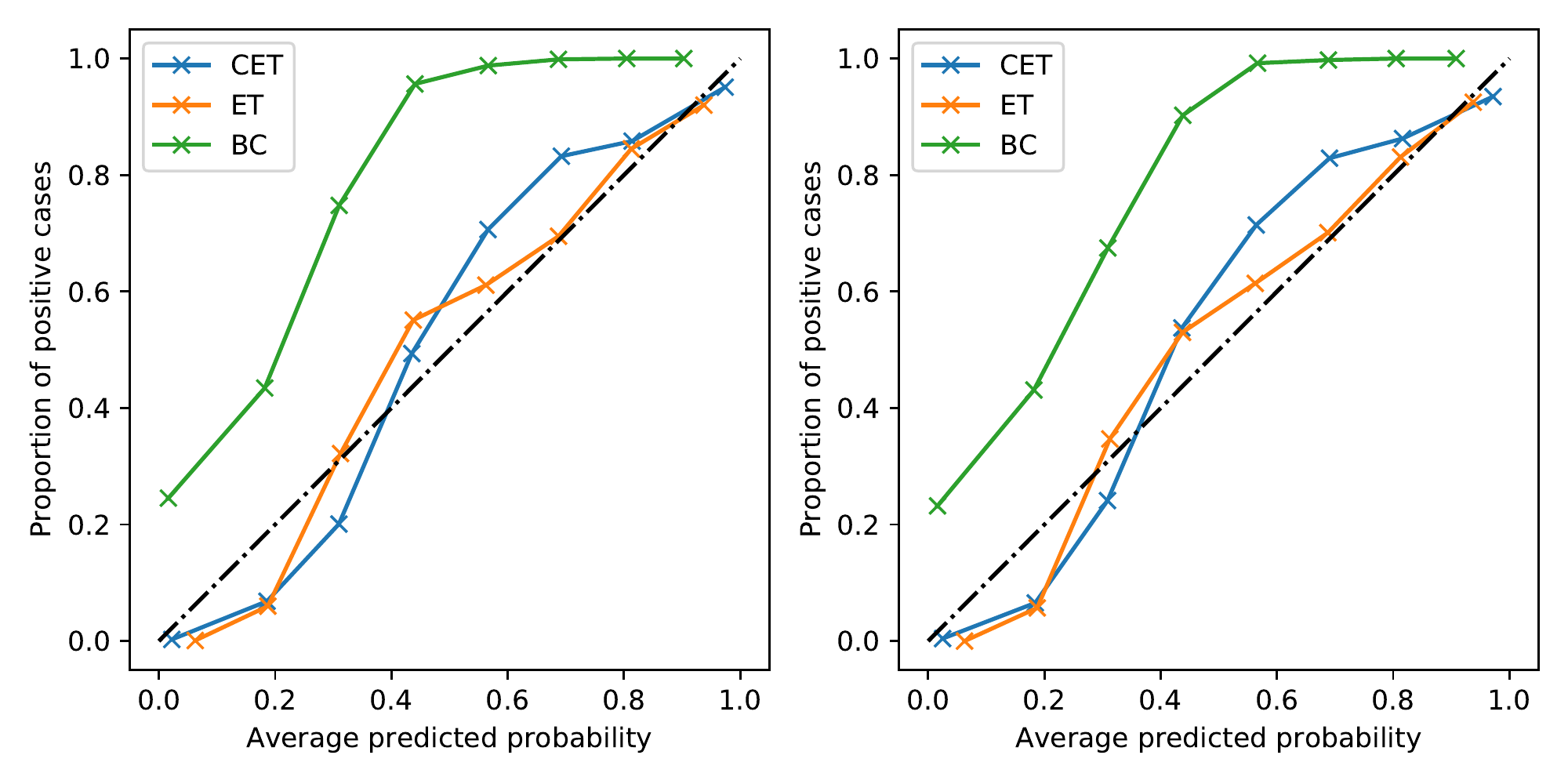}}
\caption{Calibration curves on synthetic data.}
\label{fig:calibration}
\end{center}
\vspace{-5mm}
\end{figure}

Figure \ref{fig:calibration} shows that the event probabilities predicted by CET and ET are effectively calibrated, whereas those predicted by BC are not.

Results on all datasets use the Gumbel-Softmax estimator with temperature fixed to 0.3, which was found to optimize AUC on the validation sets.

\subsection{MIMIC-III}
MIMIC-III results are consistent with the synthetic dataset: the CET model predicts event occurrence more effectively than ET or BC, and also predicts event times more accurately than ET, but with lower concordance index compared to ET (see Table 4).
However, the degree of these differences is larger than found on the synthetic data.
In particular, event time predictions made by ET are highly inaccurate, which may be due to the high variance and long tail of event times in most of the tasks.

\begin{table}[t!]
\caption{Performance metrics on MIMIC-III dataset.}
\label{mimic-performance}
\vspace{1mm}
\begin{center}
\begin{small}
\begin{sc}
\begin{tabular}{llrrr}
\toprule
& Lab & AUC & MRAE & CI\\
\midrule
\multirow{10}{*}{\rotatebox[origin=c]{90}{CET}} & CSF & \textbf{0.77}$\pm$\textbf{0.01} & \textbf{3.09}$\pm$\textbf{7.55} & 0.53$\pm$0.04\\
& Trop.  & \textbf{0.78}$\pm$\textbf{0.00} & \textbf{0.11}$\pm$\textbf{0.01} & 0.59$\pm$0.03\\
& Intub. & \textbf{0.85}$\pm$\textbf{0.00} & \textbf{0.14}$\pm$\textbf{0.01} & 0.67$\pm$0.01\\
& Pleur.   & \textbf{0.76}$\pm$\textbf{0.00} & \textbf{0.05}$\pm$\textbf{0.01} & 0.47$\pm$0.03\\
& TSH       & \textbf{0.70}$\pm$\textbf{0.00} & \textbf{0.22}$\pm$\textbf{0.14} & 0.52$\pm$0.03\\
& D-Dim.   & \textbf{0.77}$\pm$\textbf{0.00} & \textbf{0.07}$\pm$\textbf{0.03} & 0.50$\pm$0.05\\
& Urob.   & \textbf{0.76}$\pm$\textbf{0.00} & \textbf{0.13}$\pm$\textbf{0.01} & 0.62$\pm$0.03\\
& ANA       & \textbf{0.74}$\pm$\textbf{0.01} & \textbf{0.04}$\pm$\textbf{0.01} & 0.50$\pm$0.04\\
& Amm.   & \textbf{0.77}$\pm$\textbf{0.01} & \textbf{0.33}$\pm$\textbf{0.74} & 0.50$\pm$0.05\\
& Lipase    & \textbf{0.73}$\pm$\textbf{0.00} & \textbf{0.14}$\pm$\textbf{0.01} & 0.60$\pm$0.02\\
& Avg       & \textbf{0.76}$\pm$\textbf{0.00} & \textbf{0.43}$\pm$\textbf{0.78} & 0.55$\pm$0.02\\
\midrule
\multirow{10}{*}{\rotatebox[origin=c]{90}{ET}} & CSF & 0.58$\pm$0.02 & 3.1e9$\pm$5.0e9 & \textbf{0.59}$\pm$\textbf{0.02}\\
& Trop. & 0.72$\pm$0.01 & 9.1e5$\pm$1.8e6 & \textbf{0.71}$\pm$\textbf{0.01}\\
& Intub. & 0.74$\pm$0.02 & 1.7e1$\pm$5.6e0 & \textbf{0.73}$\pm$\textbf{0.02}\\
& Pleur. & 0.59$\pm$0.02 & 5.6e2$\pm$1.3e3 & \textbf{0.59}$\pm$\textbf{0.02}\\
& TSH & 0.60$\pm$0.01 & 3.1e5$\pm$3.4e5 & \textbf{0.60}$\pm$\textbf{0.01}\\
& D-Dim. & 0.60$\pm$0.02 & 3.9e3$\pm$8.7e3 & \textbf{0.56}$\pm$\textbf{0.01}\\
& Urob. & 0.72$\pm$0.01 & 8.0e3$\pm$1.0e4 & \textbf{0.68}$\pm$\textbf{0.01}\\
& ANA & 0.56$\pm$0.02 & 4.1e3$\pm$1.2e4 & \textbf{0.56}$\pm$\textbf{0.03}\\
& Amm. & 0.60$\pm$0.02 & 4.3e3$\pm$9.5e3 & \textbf{0.61}$\pm$\textbf{0.02}\\
& Lipase & 0.66$\pm$0.01 & 1.3e6$\pm$2.7e6 & \textbf{0.68}$\pm$\textbf{0.02}\\
& Avg & 0.64$\pm$0.01 & 3.1e8$\pm$5.0e8 & \textbf{0.63}$\pm$\textbf{0.01}\\
\midrule
\multirow{10}{*}{\rotatebox[origin=c]{90}{BC}} & CSF & 0.75$\pm$0.01 & &\\
& Trop. & 0.78$\pm$0.00 & &\\
& Intub. & 0.81$\pm$0.00 & &\\
& Pleur. & 0.75$\pm$0.00 & &\\
& TSH & 0.70$\pm$0.00 & &\\
& D-Dim. & 0.76$\pm$0.00 & &\\
& Urob. & 0.75$\pm$0.00 & &\\
& ANA & 0.73$\pm$0.01 & &\\
& Amm. & 0.77$\pm$0.00 & &\\
& Lipase & 0.71$\pm$0.00 & &\\
& Avg & 0.75$\pm$0.00 & &\\
\bottomrule
\end{tabular}
\end{sc}
\end{small}
\end{center}
\vspace{-4mm}
\end{table}

Direct prediction of known event occurrences yields AUCs ranging from 0.74 (TSH) to 0.90 (Intubation) with an average of 0.80. 
It is notable that performance on most tasks is high, demonstrating that important diagnostic tests ordered by care providers can be effectively predicted based on patient profiles over the first 24 hours, even when many events are censored.
Although censoring is artificial on MIMIC-III, it is natural in most medical prediction settings, wherein many patients are lost to follow-up before events of interest can be observed.
Moreover, follow-up rates are often correlated with events of interest, leading to biased results when these patients are removed from the dataset.
The CET framework allows event occurrence to be predicted effectively in all patients, not just those who have been followed for a sufficiently long period.
% AUCs are 0.80 0.82 0.90 0.79 0.74 0.79 0.83 0.77 0.81 0.78 0.80 (avg)

\subsection{Reddit}
Reddit results remain consistent with previous experiments.
Event occurrence probabilities predicted by CET are superior to those predicted by ET and BC, as measured via AUC, and event time predictions are more accurate than those predicted by ET.
On the other hand, ET orders events more effectively than CET, as measured via CI (see Table 5).
% AUCs are 0.64, 0.69, 0.67, 0.68, 0.68, 0.69, 0.67, 0.69, 0.94 (0.71 avg)

\begin{table}[t!]
\caption{Performance metrics on Reddit dataset.}
\label{reddit-performance}
\vspace{1mm}
\begin{center}
\begin{small}
\begin{sc}
\begin{tabular}{llrrr}
\toprule
& SubR & AUC & MRAE & CI\\
\midrule
\multirow{10}{*}{\rotatebox[origin=c]{90}{CET}} & ADHD & \textbf{0.63}$\pm$\textbf{0.00} & \textbf{0.04}$\pm$\textbf{0.00} & 0.58$\pm$0.01\\
& Anx. & \textbf{0.69}$\pm$\textbf{0.00} & \textbf{0.04}$\pm$\textbf{0.00} & 0.59$\pm$0.01\\
& Books & \textbf{0.67}$\pm$\textbf{0.00} & \textbf{0.06}$\pm$\textbf{0.00} & 0.59$\pm$0.00\\
& Dep. & \textbf{0.68}$\pm$\textbf{0.00} & \textbf{0.07}$\pm$\textbf{0.00} & 0.62$\pm$0.01\\
& Fit. & \textbf{0.68}$\pm$\textbf{0.00} & \textbf{0.09}$\pm$\textbf{0.00} & 0.64$\pm$0.00\\
& LPT & \textbf{0.68}$\pm$\textbf{0.00} & \textbf{0.08}$\pm$\textbf{0.00} & 0.60$\pm$0.00\\
& MH & \textbf{0.68}$\pm$\textbf{0.00} & \textbf{0.02}$\pm$\textbf{0.00} & 0.55$\pm$0.01\\
& SW & \textbf{0.68}$\pm$\textbf{0.00} & \textbf{0.03}$\pm$\textbf{0.00} & 0.57$\pm$0.01\\
& WN & \textbf{0.93}$\pm$\textbf{0.01} & \textbf{0.01}$\pm$\textbf{0.00} & 0.73$\pm$0.02\\
& Avg & \textbf{0.70}$\pm$\textbf{0.00} & \textbf{0.05}$\pm$\textbf{0.00} & 0.61$\pm$0.01\\
\midrule
\multirow{10}{*}{\rotatebox[origin=c]{90}{ET}} & ADHD & 0.59$\pm$0.01 & 3.74$\pm$0.41 & \textbf{0.64}$\pm$\textbf{0.01}\\
& Anx. & 0.68$\pm$0.00 & 3.37$\pm$0.32 & \textbf{0.72}$\pm$\textbf{0.00}\\
& Books & 0.66$\pm$0.00 & 2.50$\pm$0.36 & \textbf{0.67}$\pm$\textbf{0.00}\\
& Dep. & 0.67$\pm$0.00 & 1.86$\pm$0.19 & \textbf{0.70}$\pm$\textbf{0.00}\\
& Fit. & 0.67$\pm$0.00 & 0.95$\pm$0.10 & \textbf{0.66}$\pm$\textbf{0.00}\\
& LPT & 0.68$\pm$0.00 & 1.36$\pm$0.26 & \textbf{0.65}$\pm$\textbf{0.00}\\
& MH & 0.63$\pm$0.00 & 6.45$\pm$0.96 & \textbf{0.65}$\pm$\textbf{0.01}\\
& SW & 0.66$\pm$0.00 & 4.18$\pm$0.50 & \textbf{0.70}$\pm$\textbf{0.00}\\
& WN & 0.37$\pm$0.05 & 10.10$\pm$2.91 & \textbf{0.39}$\pm$\textbf{0.05}\\
& Avg & 0.62$\pm$0.01 & 3.83$\pm$0.52 & \textbf{0.65}$\pm$\textbf{0.01}\\
\midrule
\multirow{10}{*}{\rotatebox[origin=c]{90}{BC}} & ADHD & 0.62$\pm$0.00 & &\\
& Anx. & 0.68$\pm$0.00 & &\\
& Books & 0.66$\pm$0.00 & &\\
& Dep. & 0.67$\pm$0.00 & &\\
& Fit. & 0.66$\pm$0.00 & &\\
& LPT & 0.67$\pm$0.00 & &\\
& MH & 0.67$\pm$0.00 & &\\
& SW & 0.68$\pm$0.00 & &\\
& WN & 0.92$\pm$0.01 & &\\
& Avg & 0.69$\pm$0.00 & &\\
\bottomrule
\end{tabular}
\end{sc}
\end{small}
\end{center}
\vspace{-4mm}
\end{table}

Direct prediction of known event occurrences, {\em i.e.}, subreddit posts, yields AUCs ranging from 0.64 (r/ADHD) to 0.94 (r/worldnews), with an average of 0.71.
Good prediction performance, although not as high compared to MIMIC-III, suggests that Reddit users' tendency to post to specific subreddits -- including several related to mental health, ({\em e.g.}, r/ADHD, r/depression, r/mentalhealth, r/SuicideWatch), can be predicted effectively from a small number of early comments.
Prediction performance may be substantially higher when using a more sophisticated natural language model, whereas our current aim was to demonstrate the advantages of CET compared to alternative learning frameworks.

The CET model learns from censored event times to predict the probability that users will post to a given subreddit.
This is particularly advantageous when predicting mental health status, as many users with mental health problems may discontinue social media activity before they might otherwise decide to post.
Good prediction performance also suggests that CET might be effective for predicting other social media activity, or in recommender systems that predict user interest in specific media content.

\section{Conclusion}
In this work we have presented conditional event time models, argued that they are advantageous when modeling event occurrence and event times in a variety of real-world settings, and described how they can be implemented as a neural network with a binary stochastic layer representing the unknown, eventual occurrence of each event of interest.
Results demonstrate that CET yields superior event occurrence probabilities and event time predictions compared to alternative approaches across one synthetic and two real-world datasets comprising a total of 21 distinct prediction tasks.
Learning of CET models on large-scale datasets is facilitated by recent, improved methods for estimating gradients across categorical variables in neural networks.
We believe CET, rather than alternative event time models, should be preferred when learning from multiple censored events, particularly when accurate prediction of eventual event occurrence is a primary goal.
Future work will focus on evaluating CET in additional real-world settings, including prediction of medical diagnoses, wherein learning event occurrence probabilities from censored events is critical to avoid selection biases that may otherwise confound results.


\bibliography{ncetm}

\begin{thebibliography}{30}
\providecommand{\natexlab}[1]{#1}
\providecommand{\url}[1]{\texttt{#1}}
\expandafter\ifx\csname urlstyle\endcsname\relax
  \providecommand{\doi}[1]{doi: #1}\else
  \providecommand{\doi}{doi: \begingroup \urlstyle{rm}\Url}\fi

\bibitem[Abadi et~al.(2016)Abadi, Barham, Chen, Chen, Davis, Dean, Devin,
  Ghemawat, Irving, Isard, et~al.]{abadi2016tensorflow}
Abadi, M., Barham, P., Chen, J., Chen, Z., Davis, A., Dean, J., Devin, M.,
  Ghemawat, S., Irving, G., Isard, M., et~al.
\newblock Tensorflow: A system for large-scale machine learning.
\newblock In \emph{12th $\{$USENIX$\}$ Symposium on Operating Systems Design
  and Implementation ($\{$OSDI$\}$ 16)}, pp.\  265--283, 2016.

\bibitem[Cer et~al.(2018)Cer, Yang, Kong, Hua, Limtiaco, John, Constant,
  Guajardo-Cespedes, Yuan, Tar, et~al.]{cer2018universal}
Cer, D., Yang, Y., Kong, S.-y., Hua, N., Limtiaco, N., John, R.~S., Constant,
  N., Guajardo-Cespedes, M., Yuan, S., Tar, C., et~al.
\newblock Universal sentence encoder.
\newblock \emph{arXiv preprint arXiv:1803.11175}, 2018.

\bibitem[Chapfuwa et~al.(2018)Chapfuwa, Tao, Li, Page, Goldstein, Duke, and
  Henao]{chapfuwa2018adversarial}
Chapfuwa, P., Tao, C., Li, C., Page, C., Goldstein, B., Duke, L.~C., and Henao,
  R.
\newblock Adversarial time-to-event modeling.
\newblock In \emph{International Conference on Machine Learning}, pp.\
  735--744, 2018.

\bibitem[Cox(1972)]{cox1972regression}
Cox, D.~R.
\newblock Regression models and life-tables.
\newblock \emph{Journal of the Royal Statistical Society: Series B
  (Methodological)}, 34\penalty0 (2):\penalty0 187--202, 1972.

\bibitem[Dovidio \& Fiske(2012)Dovidio and Fiske]{dovidio2012under}
Dovidio, J.~F. and Fiske, S.~T.
\newblock Under the radar: how unexamined biases in decision-making processes
  in clinical interactions can contribute to health care disparities.
\newblock \emph{American journal of public health}, 102\penalty0 (5):\penalty0
  945--952, 2012.

\bibitem[Elandt-Johnson(1976)]{elandt-johnson_conditional_1976}
Elandt-Johnson, R.~C.
\newblock Conditional failure time distributions under competing risk theory
  with dependent failure times and proportional hazard rates.
\newblock \emph{Scandinavian Actuarial Journal}, 1976\penalty0 (1):\penalty0
  37--51, January 1976.
\newblock ISSN 0346-1238.
\newblock \doi{10.1080/03461238.1976.10405934}.

\bibitem[Farewell(1982)]{farewell_use_1982}
Farewell, V.~T.
\newblock The {Use} of {Mixture} {Models} for the {Analysis} of {Survival}
  {Data} with {Long}-{Term} {Survivors}.
\newblock \emph{Biometrics}, 38\penalty0 (4):\penalty0 1041--1046, 1982.
\newblock ISSN 0006-341X.
\newblock \doi{10.2307/2529885}.

\bibitem[Gaynor et~al.(1993)Gaynor, Feuer, Tan, Wu, Little, Straus, Clarkson,
  and Brennan]{gaynor_use_1993}
Gaynor, J.~J., Feuer, E.~J., Tan, C.~C., Wu, D.~H., Little, C.~R., Straus,
  D.~J., Clarkson, B.~D., and Brennan, M.~F.
\newblock On the {Use} of {Cause}-{Specific} {Failure} and {Conditional}
  {Failure} {Probabilities}: {Examples} {From} {Clinical} {Oncology} {Data}.
\newblock \emph{Journal of the American Statistical Association}, 88\penalty0
  (422):\penalty0 400--409, 1993.
\newblock ISSN 0162-1459.
\newblock \doi{10.2307/2290318}.

\bibitem[Grathwohl et~al.(2017)Grathwohl, Choi, Wu, Roeder, and
  Duvenaud]{grathwohl2017backpropagation}
Grathwohl, W., Choi, D., Wu, Y., Roeder, G., and Duvenaud, D.
\newblock Backpropagation through the void: Optimizing control variates for
  black-box gradient estimation, 2017.

\bibitem[Harrell~Jr et~al.(1984)Harrell~Jr, Lee, Califf, Pryor, and
  Rosati]{harrell1984regression}
Harrell~Jr, F.~E., Lee, K.~L., Califf, R.~M., Pryor, D.~B., and Rosati, R.~A.
\newblock Regression modelling strategies for improved prognostic prediction.
\newblock \emph{Statistics in medicine}, 3\penalty0 (2):\penalty0 143--152,
  1984.

\bibitem[Jang et~al.(2016)Jang, Gu, and Poole]{jang2016categorical}
Jang, E., Gu, S., and Poole, B.
\newblock Categorical reparameterization with gumbel-softmax.
\newblock \emph{arXiv preprint arXiv:1611.01144}, 2016.

\bibitem[Johnson et~al.(2016)Johnson, Pollard, Shen, Li-wei, Feng, Ghassemi,
  Moody, Szolovits, Celi, and Mark]{johnson2016mimic}
Johnson, A.~E., Pollard, T.~J., Shen, L., Li-wei, H.~L., Feng, M., Ghassemi,
  M., Moody, B., Szolovits, P., Celi, L.~A., and Mark, R.~G.
\newblock Mimic-iii, a freely accessible critical care database.
\newblock \emph{Scientific data}, 3:\penalty0 160035, 2016.

\bibitem[Kalbfleisch \& Prentice(2011)Kalbfleisch and
  Prentice]{kalbfleisch2011statistical}
Kalbfleisch, J.~D. and Prentice, R.~L.
\newblock \emph{The statistical analysis of failure time data}, volume 360.
\newblock John Wiley \& Sons, 2011.

\bibitem[Katzman et~al.(2018{\natexlab{a}})Katzman, Shaham, Cloninger, Bates,
  Jiang, and Kluger]{katzman2018deepsurv}
Katzman, J.~L., Shaham, U., Cloninger, A., Bates, J., Jiang, T., and Kluger, Y.
\newblock Deepsurv: personalized treatment recommender system using a cox
  proportional hazards deep neural network.
\newblock \emph{BMC medical research methodology}, 18\penalty0 (1):\penalty0
  24, 2018{\natexlab{a}}.

\bibitem[Katzman et~al.(2018{\natexlab{b}})Katzman, Shaham, Cloninger, Bates,
  Jiang, and Kluger]{katzman_deepsurv:_2018}
Katzman, J.~L., Shaham, U., Cloninger, A., Bates, J., Jiang, T., and Kluger, Y.
\newblock {DeepSurv}: personalized treatment recommender system using a {Cox}
  proportional hazards deep neural network.
\newblock \emph{BMC Medical Research Methodology}, 18\penalty0 (1):\penalty0
  24, December 2018{\natexlab{b}}.
\newblock ISSN 1471-2288.
\newblock \doi{10.1186/s12874-018-0482-1}.

\bibitem[Kingma \& Ba(2014)Kingma and Ba]{kingma2014adam}
Kingma, D.~P. and Ba, J.
\newblock Adam: A method for stochastic optimization.
\newblock \emph{arXiv preprint arXiv:1412.6980}, 2014.

\bibitem[Kvamme et~al.(2019{\natexlab{a}})Kvamme, Borgan, and
  Scheel]{kvamme2019time}
Kvamme, H., Borgan, O., and Scheel, I.
\newblock Time-to-event prediction with neural networks and cox regression.
\newblock \emph{Journal of Machine Learning Research}, 20\penalty0
  (129):\penalty0 1--30, 2019{\natexlab{a}}.

\bibitem[Kvamme et~al.(2019{\natexlab{b}})Kvamme, Borgan, and
  Scheel]{kvamme_time--event_2019}
Kvamme, H., Borgan, O., and Scheel, I.
\newblock Time-to-{Event} {Prediction} with {Neural} {Networks} and {Cox}
  {Regression}.
\newblock \emph{arXiv:1907.00825 [cs, stat]}, September 2019{\natexlab{b}}.
\newblock arXiv: 1907.00825.

\bibitem[Lee et~al.(2018)Lee, Zame, Yoon, and van~der Schaar]{lee2018deephit}
Lee, C., Zame, W.~R., Yoon, J., and van~der Schaar, M.
\newblock Deephit: A deep learning approach to survival analysis with competing
  risks.
\newblock In \emph{Thirty-Second AAAI Conference on Artificial Intelligence},
  2018.

\bibitem[Maddison et~al.(2016)Maddison, Mnih, and Teh]{maddison2016concrete}
Maddison, C.~J., Mnih, A., and Teh, Y.~W.
\newblock The concrete distribution: A continuous relaxation of discrete random
  variables.
\newblock \emph{arXiv preprint arXiv:1611.00712}, 2016.

\bibitem[Pardes(2018)]{reddit_wired}
Pardes, A.
\newblock The {Inside} {Story} of {Reddit}'s {Redesign}.
\newblock \emph{Wired}, 2018.
\newblock ISSN 1059-1028.

\bibitem[Ranganath et~al.(2016)Ranganath, Perotte, Elhadad, and
  Blei]{ranganath2016deep}
Ranganath, R., Perotte, A., Elhadad, N., and Blei, D.
\newblock Deep survival analysis.
\newblock \emph{arXiv preprint arXiv:1608.02158}, 2016.

\bibitem[Ren et~al.(2019)Ren, Qin, Zheng, Yang, Zhang, Qiu, and
  Yu]{ren_deep_2018}
Ren, K., Qin, J., Zheng, L., Yang, Z., Zhang, W., Qiu, L., and Yu, Y.
\newblock Deep recurrent survival analysis.
\newblock In \emph{Proceedings of the AAAI Conference on Artificial
  Intelligence}, volume~33, pp.\  4798--4805, 2019.

\bibitem[Shen et~al.(2018)Shen, Wang, Wang, Min, Su, Zhang, Li, Henao, and
  Carin]{shen2018baseline}
Shen, D., Wang, G., Wang, W., Min, M.~R., Su, Q., Zhang, Y., Li, C., Henao, R.,
  and Carin, L.
\newblock Baseline needs more love: On simple word-embedding-based models and
  associated pooling mechanisms.
\newblock \emph{arXiv preprint arXiv:1805.09843}, 2018.

\bibitem[Tucker et~al.(2017)Tucker, Mnih, Maddison, Lawson, and
  Sohl-Dickstein]{tucker2017rebar}
Tucker, G., Mnih, A., Maddison, C.~J., Lawson, J., and Sohl-Dickstein, J.
\newblock Rebar: Low-variance, unbiased gradient estimates for discrete latent
  variable models.
\newblock In \emph{Advances in Neural Information Processing Systems}, pp.\
  2627--2636, 2017.

\bibitem[von Allmen et~al.(2015)von Allmen, Weiss, Tevaearai, Kuemmerli,
  Tinner, Carrel, Schmidli, and Dick]{von2015completeness}
von Allmen, R.~S., Weiss, S., Tevaearai, H.~T., Kuemmerli, C., Tinner, C.,
  Carrel, T.~P., Schmidli, J., and Dick, F.
\newblock Completeness of follow-up determines validity of study findings:
  results of a prospective repeated measures cohort study.
\newblock \emph{PLoS One}, 10\penalty0 (10), 2015.

\bibitem[Wei(1992)]{wei1992accelerated}
Wei, L.-J.
\newblock The accelerated failure time model: a useful alternative to the cox
  regression model in survival analysis.
\newblock \emph{Statistics in medicine}, 11\penalty0 (14-15):\penalty0
  1871--1879, 1992.

\bibitem[Williams(1992)]{williams1992simple}
Williams, R.~J.
\newblock Simple statistical gradient-following algorithms for connectionist
  reinforcement learning.
\newblock \emph{Machine learning}, 8\penalty0 (3-4):\penalty0 229--256, 1992.

\bibitem[Yin \& Zhou(2019)Yin and Zhou]{yin2018arm}
Yin, M. and Zhou, M.
\newblock {ARM}: Augment-{REINFORCE}-merge gradient for stochastic binary
  networks.
\newblock In \emph{International Conference on Learning Representations}, 2019.

\bibitem[Zheng et~al.(2019)Zheng, Yuan, and Wu]{zheng_safe:_2019}
Zheng, P., Yuan, S., and Wu, X.
\newblock {SAFE}: {A} {Neural} {Survival} {Analysis} {Model} for {Fraud}
  {Early} {Detection}.
\newblock \emph{Proceedings of the AAAI Conference on Artificial Intelligence},
  33:\penalty0 1278--1285, July 2019.
\newblock ISSN 2374-3468, 2159-5399.
\newblock \doi{10.1609/aaai.v33i01.33011278}.

\end{thebibliography}
\bibliographystyle{icml2020}

\cleardoublepage

\appendix

\section*{Appendix A: Derivation of Equation (8) (Section 3.2)}

From equation (7) of section 3.2, we have the following expression for $p_\theta(t_i, s_i | c_i, \bm{x}_i)$:
\begin{align}
\label{eq:1}
    p_\theta(t_i, s_i | c_i, \bm{x}_i) = & \ p_\theta(t_i, s_i=1 | c_i=1, \bm{x}_i)^{s_ic_i} \\
    & \times p_\theta(t_i, s_i=0 | c_i=1, \bm{x}_i)^{(1-s_i)c_i} \notag \\
    & \times p(t_i, s_i=1 | c_i=0, \bm{x}_i)^{s_i(1-c_i)} \notag \\
    & \times p(t_i, s_i=0 | c_i=1, \bm{x}_i)^{(1-s_i)(1-c_i)} . \notag
\end{align}

We assign a small probability $0<\epsilon\ll1$ to $P(\mathcal{F}_i < \infty |c_i=0)$, so that:
\begin{align}
p(t_i, s_i=1 | c_i=0, \bm{x}_i) = & \epsilon f_\theta(t_i| \bm{x}_i)G_i(t_i) \\
p(t_i, s_i=0 | c_i=0, \bm{x}_i) = & (1-\epsilon)g_i(t_i) \notag \\
& + \epsilon g_i(t_i)F_\theta(t_i | \bm{x}_i) \\
\approx & g_i(t_i) \notag .
\end{align}

This allows us to expand (\ref{eq:1}):
\begin{align}
    p_\theta(t_i, s_i | c_i, \bm{x}_i) = & \big(f_\theta(t_i| \bm{x}_i)G_i(t_i)\big)^{s_i c_i} \\
    & \times \big(g_i(t_i)F_\theta(t_i| \bm{x}_i)\big)^{(1-s_i) c_i} \notag \\
    & \times \big(\epsilon f_\theta(t_i| \bm{x}_i)G_i(t_i)\big)^{s_i (1-c_i)} \notag \\
    & \times g_i(t_i)^{(1-s_i)(1-c_i)}. \notag
\end{align}

Simplifying, we obtain:
\begin{align}
    p_\theta(t_i, s_i | c_i, \bm{x}_i) = & \big(f_\theta(t_i| \bm{x}_i)G_i(t_i)\big)^{s_i} \\
    & \times F_\theta(t_i| \bm{x}_i)^{(1-s_i) c_i} \notag \\
    & \times g_i(t_i)^{(1-s_i)}, \notag \\
    & \times \epsilon^{s_i(1-c_i)}
\end{align}

We then remove terms that do not depend on $\theta$ or $c_i$, including $g_i(\cdot)$ and $G_i(\cdot)$, to obtain equation (8) from section 3.2:
\begin{align}
\begin{aligned}
    p_\theta(t_i, s_i|c_i, \bm{x}_i) \approxprop & \ \epsilon^{s_i(1-c_i)} \\
    & \times f_\theta(t_i|\bm{x}_i)^{s_i}F_\theta(t_i|\bm{x}_i)^{(1-s_i)c_i} .
\end{aligned}
\end{align}

\section*{Appendix B: Descriptive Statistics, MIMIC-III}

MIMIC-III may be accessed, following approval, at \url{https://mimic.physionet.org}. A complete description of this dataset, including descriptive statistics for all tables used in this work, may be found in \citep{johnson2016mimic}.

\section*{Appendix C: Descriptive Statistics, Reddit}

Reddit data was accessed via the pushshift.io API. Code needed to generate our dataset is available at \url{http://anon.site}.

Our final dataset included the earliest 20 comments and first subreddit submissions to each of the nine chosen subreddits from 492,059 unique Reddit users active between 2005 and 2020. Supplementary table (\ref{tab:reddit}) shows the breakdown of comments and submissions by year:

\begin{table}[h]
    \begin{center}
    \begin{small}
    \begin{sc}
    \begin{tabular}{lrr}
        \toprule
        Year & Submissions & Comments\\
        \midrule
        2005&0&69\\
        2006&0&5228\\
        2007&0&18118\\
        2008&136&42624\\
        2009&985&123788\\
        2010&5285&301978\\
        2011&14294&748479\\
        2012&30456&1254574\\
        2013&43994&1339624\\
        2014&57023&1471171\\
        2015&81687&1559287\\
        2016&93793&1613979\\
        2017&105302&1362261\\
        2018&89941&0\\
        2019&58704&0\\
        2020&3717&0\\
        \bottomrule
    \end{tabular}
    \end{sc}
    \end{small}
    \end{center}
    \caption{Comments and submissions by year in Reddit dataset}
    \label{tab:reddit}
\end{table}

\begin{figure}[h]
\begin{center}
\includegraphics[]{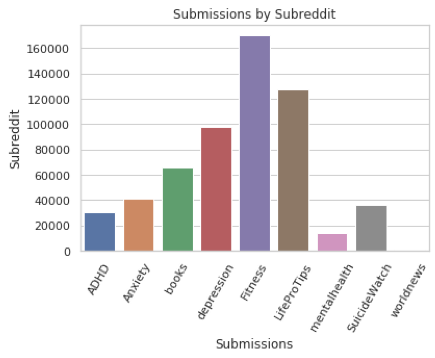}
\caption{Total submissions to each subreddit}
\label{fig:reddit_posts}
\end{center}
\end{figure}

Supplementary figure (\ref{fig:reddit_posts}) shows the number of users who posted to each subreddit. Supplementary table (\ref{tab:numposts}) shows that the majority of users posted to only one of the nine subreddits, and none posted to eight or all nine.

\begin{table}[h]
    \begin{center}
    \begin{small}
    \begin{sc}
    \begin{tabular}{ll}
        \toprule
        Num. subreddits & Num. users\\
        \midrule
        1&416965\\
        2&60795\\
        3&11236\\
        4&2408\\
        5&525\\
        6&113\\
        7&17\\
        8&0\\
        9&0\\
        \bottomrule
    \end{tabular}
    \end{sc}
    \end{small}
    \end{center}
    \caption{Reddit users by the number of distinct subreddits to which they posted}
    \label{tab:numposts}
\end{table}

\section*{Appendix D: Additional Experiment Details}

All models were trained in Tensorflow 1.10 \citep{abadi2016tensorflow} using a single NVIDIA Titan XP GPU.

Hyperparameters were explored via random search, selected uniformly in the ranges listed in supplementary table (\ref{tab:hyp}), and tuned to optimize AUC of the CET model on the validation set.

\begin{table}[h]
    \begin{center}
    \begin{small}
    \begin{sc}
    \begin{tabular}{ll}
        \toprule
        Hyperparameter & Range\\
        \midrule
        Estimator & \{Gumbel-Softmax, ARM\}\\
        Num. $c_i$ samples & $\{30, ..., 200\}$\\
        $\log \epsilon$ & $(-4, 0)$\\
        Hidden units & $\{100, ..., 1000\}$\\
        Gumbel-SM Temp. & $(0, 1)$\\
    \end{tabular}
    \end{sc}
    \end{small}
    \end{center}
    \caption{Hyperparameter ranges for random search}
    \label{tab:hyp}
\end{table}

The Gumbel-Softmax estimator with a temperature of approximately .3 and was found to be optimal on all three datasets. 100 samples were adequate on all datasets; further increasing the number of samples did not improve performance. Optimal values of $\log \epsilon$ were approximately -2 on all datasets. Layer widths of 750 (for $h_\phi(\bm{x}_i)$, $\mu_\theta(\bm{x}_i, \bm{c}_i)$, and $\nu_\theta(\bm{x}_i, \bm{c}_i)$) were used in the final MIMIC-III and Reddit models, whereas widths of 100 were used in the final Synthetic model.

\end{document}